%% file: acl_latex.tex
\pdfoutput=1

\documentclass[11pt]{article}

\usepackage[final]{acl}

\usepackage{times}
\usepackage{latexsym}
\usepackage{cleveref}
\usepackage{booktabs}
\usepackage{multirow}
\usepackage[T1]{fontenc}
\usepackage[utf8]{inputenc}
\usepackage{microtype}
\usepackage{inconsolata}
\usepackage{amssymb}    
\usepackage{pifont}     
\usepackage[framemethod=TikZ]{mdframed}
\usepackage{graphicx}
\usepackage[inkscapelatex=false]{svg}
\usepackage{subcaption}
\usepackage{enumitem}

\usepackage[T1]{fontenc}

\usepackage[utf8]{inputenc}

\usepackage{microtype}

\usepackage{inconsolata}

\usepackage{graphicx}

\title{LLMs Can Compensate for Deficiencies in Visual Representations}

\author{
    \textbf{Sho Takishita}$^{1,2}$\thanks{Equal contribution}\thanks{Research conducted during a research stay at MBZUAI.}
    \quad
    \textbf{Jay Gala}$^{2*}$
    \quad
    \textbf{Abdelrahman Mohamed}$^{2}$
    \quad
    \vspace{2pt} \\
    \textbf{Kentaro Inui}$^{2,3,4}$
    \quad
    \textbf{Yova Kementchedjhieva}$^{2}$
    \quad
    \vspace{5pt} \\
    $^{1}$Fujitsu Limited
    \quad
    $^{2}$MBZUAI
    \quad
    $^{3}$Tohoku University
    \quad
    $^{4}$RIKEN
    \quad
    \vspace{5pt} \\
    \texttt{sho.takishita@jp.fujitsu.com} 
    \vspace{2pt} \\
    \texttt{\{jay.gala, abdelrahman.mohamed, yova.kementchedjhieva\}@mbzuai.ac.ae} 
    \vspace{2pt} \\
    \texttt{kentaro.inui@tohoku.ac.jp}
}

\begin{document}
\maketitle
\input{main}

\bibliography{custom}

\appendix
\input{appendix}

\end{document}

%% file: main.tex
\begin{abstract}
Many vision-language models (VLMs) that prove very effective at a range of multimodal tasks build on CLIP-based vision encoders, which are known to have various limitations. We investigate the hypothesis that the strong language backbone in VLMs compensates for possibly weak visual features by contextualizing or enriching them.
Using three CLIP-based VLMs, we perform controlled self-attention ablations on a carefully designed probing task. 
Our findings show that despite known limitations, CLIP visual representations offer ready-to-read semantic information to the language decoder. However, in scenarios of reduced contextualization in the visual representations, the language decoder can largely compensate for the deficiency and recover performance. This suggests a dynamic division of labor in VLMs and motivates future architectures that offload more visual processing to the language decoder.
\end{abstract}


\section{Introduction }

Vision-language models (VLMs) have made remarkable progress in recent years, with systems like MAGMA \citep{eichenberg2021magma}, BLIP-2 \citep{li2023blip020}, LLaVA \cite{liu2023visualinstructiontuning,liu2024improvedbaselinesvisualinstruction}, and Prismatic \citep{karamcheti2024prismatic} demonstrating strong performance for their time on key multimodal benchmarks. A common design choice across these models is the use of a frozen pretrained vision encoder, often based on CLIP \cite{radford2021learningtransferablevisualmodels}, paired with a pre-trained language model, which is finetuned to map visual features into text.\footnote{The vision encoder and language decoder are commonly connected with a linear projection or a multi-layer perceptron, which has largely been shown to serve the technical role of mapping between dimensionalities rather than semantic spaces \citep{schwettmann2023multimodalneuronspretrainedtextonly, verma-etal-2024-cross}.}

Despite the widespread use of CLIP, its limitations as a visual backbone are well-documented. 
CLIP representations have been shown to prioritize global over local features \cite{10.1007/978-3-031-72664-4_18}, to perform poorly in distinguishing objects which share high-level features \cite{shao2023investigatinglimitationclipmodels}, to lack fine-grained compositionality \cite{lewis2024clip}, and to exhibit quantity and size biases \cite{zhang-etal-2024-clip, abbasi2025analyzingclipsperformancelimitations}. Many of these limitations have been attributed to the contrastive training objective that CLIP employs.

Nonetheless, many modern VLMs that rely on the CLIP vision encoder perform surprisingly well, even on tasks requiring detailed visual understanding \cite{fu2023mme0, yue2024mmmu0pro0,OnoeDocci2024}. This raises a fundamental question: How do VLMs overcome the known limitations of CLIP's representations? One plausible hypothesis is that the language decoder--often much larger than the vision encoder and trained with rich linguistic supervision--plays a compensatory role, enriching or contextualizing the visual representations it receives. If true, this would suggest a more dynamic division of labor between vision and language components than currently believed.

In this work, we investigate this hypothesis through a series of controlled self-attention blocking experiments \citep{geva2023dissecting} on three VLMs, each of which pairs CLIP with a distinct language model. We ask whether the language decoder in a VLM contributes to the enrichment of image features. As a diagnostic task, we focus on the identification of localized object \textit{parts} (e.g., the \textit{ear} of a cat or the \textit{stem} of an apple). This fine-grained task requires extensive contextualization of the part region into the broader context of the object.

We build a probe using segmentation annotations from the Panoptic Parts dataset \citep{degeus2021panopticparts, meletis2020panopticparts}, and apply Logit Lens analysis \citep{nostalgebraist2020logitlens} to inspect the intermediate representations across layers of the VLM. We iteratively block self-attention in the vision encoder, language decoder, or both, and evaluate the VLM's ability to maintain the identifiability of object parts in the relevant regions.

Our results yield three key findings: \textbf{(1)} When the decoder self-attention is disabled, part identifiability is largely preserved, suggesting that the self-attention in the decoder does not substantially enrich already-good visual features.
\textbf{(2)} When both encoder and decoder self-attentions are blocked, part identifiability collapses, showing the importance of some form of context-based feature construction.
\textbf{(3)} Crucially, when only the encoder self-attention is blocked, part identifiability largely recovers--indicating that the decoder can compensate for deficiencies in visual representations when the need arises through contextualization.

These findings challenge the assumption that visual semantic understanding is fully localized in the vision encoder of VLMs. Instead, they suggest an adaptive joint processing of image inputs by VLM components, where the language decoder can step in to compensate for a degraded input from the vision encoder. This has implications for future VLM architectures, which could actively offload more of the vision processing onto the language decoder, reducing the contribution of the vision encoder to only that which the decoder cannot recover.


\section{Related Work }

Efforts in interpreting language models have led to the development of techniques that probe internal representations and decompose the mechanism behind next-token prediction. Methods such as neuron attribution \cite{dai-etal-2022-knowledge}, causal tracing \cite{meng2022locating}, attention knockout \cite{geva2023dissecting}, and logit lens \cite{nostalgebraist2020logitlens} have improved our understanding of how information flows across tokens and model layers. These have also been extended to VLMs to understand how visual and linguistic information interact.

\citet{basu2024understanding} applied causal tracing to VLMs and found that LLaVA \cite{liu2023visualinstructiontuning} retrieves information from earlier layers in the language decoder compared to unimodal language models. Building on this, \citet{jiang2024devilsmiddlelayerslarge} used attention knockout and logit lens analysis to further decompose information flow in VLMs, revealing a two-stage process: visual enrichment, where image features transfer to object tokens, and semantic refinement, where these features are interpreted through language. Similarly, \citet{zhang2024cross} demonstrated that early layers transfer global visual information into question token representations, mid-layers inject question-relevant visual features into corresponding text positions, and later layers propagate this fused representation to the last token position for answer prediction. Contrary to such grounding evidence, \citet{liu2024paying} highlighted the dominant role of language priors in VLM's outputs, while \citet{stan2024lvlmintrepret} found that in different settings LLaVA either over-relies on text input or, conversely, exhibits strong visual grounding.

While the studies above explore how VLMs \textit{generate} text from either visual input or a mix of visual and text input, there have been very few works that attempt to dissect the language model representation of the visual tokens as such. 
\citet{schwettmann2023multimodal} identified multimodal neurons in the language decoder that map visual inputs to corresponding linguistic concepts, highlighting that linear projections in VLMs lack semantics on their own when interpreted using language vocabulary. Most related to our focus, \citet{neo2025towards} used visual token ablation and logit lens analysis to reveal that LLaVA localizes object-specific information at the corresponding positional tokens and can align these with fine-grained semantic concepts beyond surface-level object categories. 

Zooming in on the vision encoder itself,  \citet{gandelsman2024interpreting,gandelsman2025interpreting} decomposed CLIP’s visual embeddings into sparse text concepts and found that different attention heads specialize in distinct semantic properties (e.g., color, shape, location, etc). Despite CLIP's strong zero-shot classification performance, various studies have highlighted its limitations, ranging from poor compositional reasoning and limited sensitivity to fine visual distinctions \citep{10.1007/978-3-031-72664-4_18,shao2023investigatinglimitationclipmodels,lewis2024clip,zhang-etal-2024-clip}, to biases related to object quantity and size \cite{abbasi2025analyzingclipsperformancelimitations}. 

Meanwhile, \citet{li2024exploring} showed that some perceived limitations in the CLIP visual representations in fact stem from the global pooling of features in similarity-based analysis and from its weak text encoder.  
In a similar vein, \citet{lin2024evaluating} noted the inability of image-text metrics like CLIPScore \citep{hessel2021clipscore} to distinguish differences in object relations, attributes, or logical structure, often treating captions as a ``bag of words.''

\begin{figure*}[t]
  \centering
  \includegraphics[width=\textwidth]{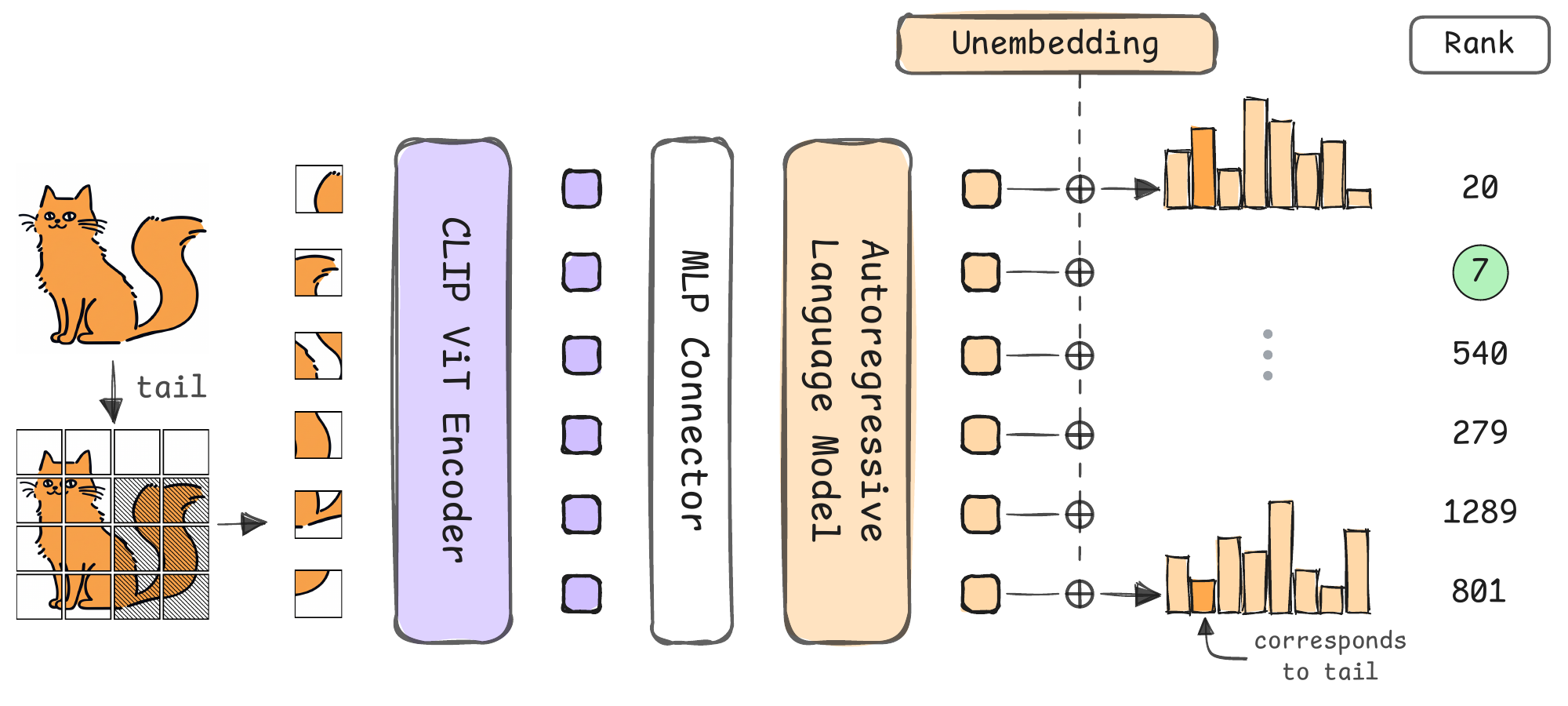}
  \caption{Overview of Object Part Identification. 
  Given an object part (e.g., the tail of a cat), localized to a region of image patches through a segmentation mask, a representation is obtained from the VLM for all relevant patches, a probability distribution is induced through LogitLens, and an identifiability rank is extracted from each distribution for the relevant label (tail). The highest rank across patches indicates the overall part identifiability.}
  \label{fig:overview}
\end{figure*}

Together, these studies tell a nuanced story of the mechanisms behind building visual representations, with their perceived and real limitations, and translating those into text. Our work investigates how visual representations evolve through the layers of the language decoder and to what extent their deficiencies are resolved in the process.


\section{Object Part Identification (OPI)}
\label{sec:OPI}

To investigate how vision-language models contextualize visual inputs, we propose a fine-grained probing task that requires extensive contextualization: identifying object parts in patch-level localized image regions. The complexity of this task arises from the fact that parts can often be overlooked in favor of the whole, while also being potentially unidentifiable in isolation of the object they belong to. The details of the probing task are described below and are visualized in Figure~\ref{fig:overview}.

\paragraph{Task.} 
OPI assumes the availability of patch-level masks which localize object parts in an input image. A single patch can contain multiple parts and a single part can span multiple patches. Probing the VLM consists of: (1) inducing a probability distribution from a VLM for every image patch, (2) extracting the rank of all part labels relevant to a patch, and (3) aggregating the ranks across all patches in the part-relevant. This results in one value per part which indicates how well the model identifies this part in this image. 

\paragraph{Dataset.} 
We derive the necessary annotations from the Pascal Panoptic Parts (Pascal-PP) dataset \citep{degeus2021panopticparts, meletis2020panopticparts} which provides high-quality pixel-level segmentation annotations for 194 part classes spanning 16 object classes, e.g., dog-head, horse-tail, person-leg, etc. 
We focus on 7 out of the 16 object classes available in Pascal-PP, specifically animal objects. This and other measures taken to reduce noise are outlined in \Cref{app:dataset_filtering}. 
After filtering, we sample up to 100 images per class, where available.

Finally, we convert pixel-level segmentation masks into patch-level masks, effectively reducing the resolution of the masks. The final dataset contains 567 unique images and 2840 unique part regions. (see \Cref{tab:dataset_stats} for more details.)

\paragraph{Probing the VLM.} 
Given a VLM,\footnote{One that follows the standard patch-level decoder prefixing method of cross-modal fusion \cite{liu2023visualinstructiontuning}.} we feed an input image all the way through and use Logit Lens  \cite{nostalgebraist2020logitlens} to analyze the semantics of the patch representations as they pass through the decoder. 
Specifically, the hidden representation of  patch $i$, $h_i \in \mathbb{R}^{d_{decoder}}$, is projected into the output vocabulary space as follows:

\begin{equation}
    p_i = \mathrm{softmax}(U \cdot \mathrm{LayerNorm}(h_i))
\end{equation}

\noindent where $U \in \mathbb{R}^{|V|\times d}$ is the unembedding matrix that projects the \(d\)-dimensional hidden representation to a $|V|$-dimensional logit space corresponding to the vocabulary, $V$, of the language decoder.

\paragraph{Rank.} 
Prior to observing the rank of a label, the probability distribution is processed to aggregate all label aliases, e.g., ``leg'', ``legs'', ``\_leg'', ``\_legs'', under a single token, \textit{leg}, summing up all their probabilities.\footnote{The label aliases are manually defined with reference to the VLM vocabulary.}
The rank of label $l$ is obtained from the probability distribution $p_i$ as follows:

\begin{equation}
\label{eq:identifiability}
    r_i = 1 - \frac{\log\left(\mathrm{argwhere}(\mathrm{argsort}(p_i)=l)\right)}{\log\left(|V|\right)}
\end{equation}

\noindent where $\mathrm{argsort}$ sorts the logits in descending order, $\mathrm{argwhere}$ returns the index of the label and $|V|$ represents the vocabulary size.
Considering the large size of the vocabulary, we apply a logarithmic transformation to emphasize the differences among high ranks. 
This rank is computed only for the parts relevant to the patch in question, as knowledge of the localization of object parts is a given.

\paragraph{Aggregation Across Regions.}
Lastly, for every part region, which optionally spans multiple patches, we perform max pooling to obtain a final identifiability score \citep{vilas2023analyzing}.
The choice of max pooling over mean pooling is motivated by \citeauthor{neo2025towards}'s (\citeyear{neo2025towards}) finding that most of the information about an object is concentrated in only a few of the object's patches; we too observed the same. Max pooling ensures that the strongest identifiability signal is reflected in the final score.

The identifiability scores obtained in this way for the parts in a single image are then averaged across all images in the dataset and enable comparisons across the identifiability of different parts in the same experimental setting and of the same parts across different experimental settings.


\section{Probing Contextualization in VLMs }
\label{sec:method}

\subsection{Model Architecture and Details}

Our study focuses on the widely adopted LLaVA architecture \citep{liu2023visualinstructiontuning,liu2024improvedbaselinesvisualinstruction}. LLaVA is a VLM which integrates an image encoder with a language model through a trained connector module in a two-stage fine-tuning recipe, with image-text caption pairs and multimodal conversations. The image encoder processes image patches and outputs patch-level visual representations that are projected into the subspace of the language model via the connector module. These projected representations together with a representation of any prompt text constitute are contextualized through causal self-attention and used to condition the autoregressive generation of new text.

\begin{table}[t]
    \centering
    \resizebox{\linewidth}{!}{
    \begin{tabular}{l p{0.6\linewidth} r r}
        \toprule
        \textbf{Object} & \textbf{Parts} & \textbf{\(N\)} & \textbf{\(P\)} \\
        \midrule
        Bird   & Beak, Foot, Head, Leg, Neck, Tail, Torso, Wing & 93 & 394 \\
        Cat    & Ear, Eye, Head, Leg, Neck, Nose, Paw, Tail, Torso & 100 & 585 \\
        Cow    & Ear, Head, Horn, Leg, Muzzle, Neck, Tail, Torso & 56 & 193 \\
        Dog    & Ear, Eye, Head, Leg, Muzzle, Neck, Nose, Paw, Tail, Torso & 100 & 518 \\
        Horse  & Ear, Head, Hoof, Leg, Muzzle, Neck, Tail, Torso & 81 & 311 \\
        Person & Arm, Ear, Eye, Foot, Hair, Hand, Head, Leg, Mouth, Neck, Nose, Torso & 92 & 701 \\
        Sheep  & Ear, Head, Horn, Leg, Muzzle, Neck, Tail, Torso & 45 & 138 \\
        \midrule
        Total & & 567 & 2840 \\
        \bottomrule
    \end{tabular}
    }
    \caption{Summary of selected object classes with available parts. \(N\): total number of images per class and \(P\): total number of unique part regions per class.}
    \label{tab:dataset_stats}
\end{table}

We carry out experiments with LLaVA-1.5 7B and 13B\footnote{\url{https://huggingface.co/collections/llava-hf/llava-15-65f762d5b6941db5c2ba07e0}}
, BakLLaVA 7B\footnote{\url{https://huggingface.co/llava-hf/bakLlava-v1-hf}}
and TinyLLaVA\footnote{\url{https://huggingface.co/bczhou/tiny-llava-v1-hf}}
\citep{zhou2024tinyllava}, all three of which use the CLIP ViT-L/14 image encoder with 24 layers \citep{radford2021learningtransferablevisualmodels}. As decoder, LLaVA-1.5 7B/13B uses Vicuna 7B/13B \citep{vicuna2023}, BakLLaVA uses Mistral 7B v0.1 \citep{jiang2023mistral}, and TinyLLaVA uses  TinyLLaMA (1.1B) \citep{zhang2024tinyllama}. All language decoders are prompted using their standard template (``USER: \texttt{<image>}'').

This selection of models with different language backbones allows us to robustly study the role of visual input contextualization in LLMs in general. Indeed, our results show consistent trends across all VLMs, with reduced magnitude for the considerably smaller TinyLLaVA model, whose results we show \Cref{tab:llava_variants_opi_scores} in the Appendix.

\begin{figure*}[t]
    \centering
    \includegraphics[width=\textwidth]{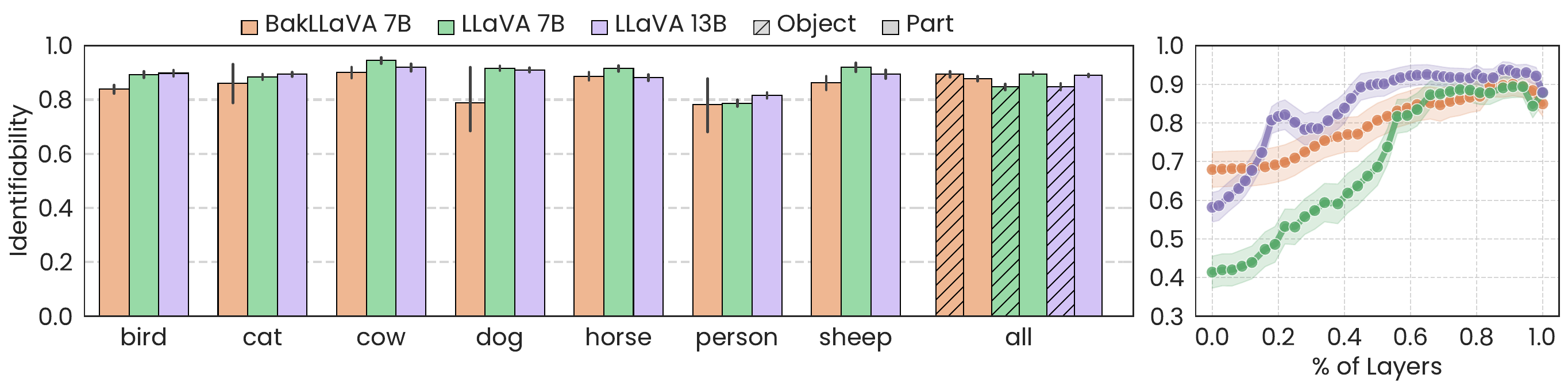}
    \caption{OPI scores for BakLLaVA 7B, LLaVA 7B, and LLaVA 13B models. Left: Per-object part identifiability with aggregated part and object-level scores at the end. Right: Layer-wise evolution of per-object part identifiability with normalized across layers as percentages for consistency.}
  \label{fig:llava_identify}
\end{figure*}

\begin{figure}[t]
  \centering
  \includegraphics[width=\columnwidth]{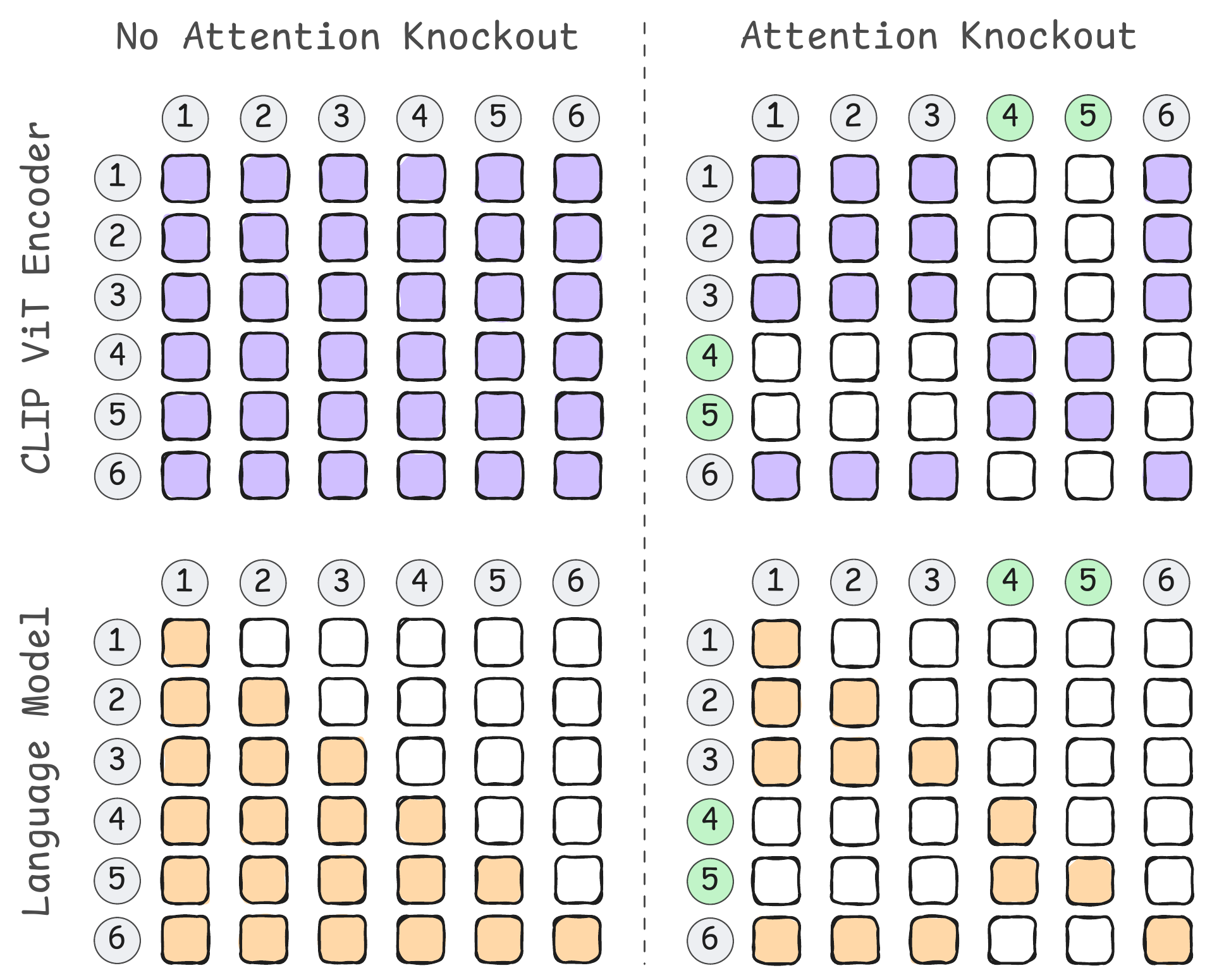}
  \caption{Attention maps from the vision encoder and language decoder with and without attention knockout. Green indicates target patches under knockout.}
  \label{fig:vlm-attention-map}
\end{figure}

\subsection{Establishing a Baseline} \label{sec:baseline}

The primary goal of this study is to understand the role and capabilities of language decoders in VLMs through the lens of how they solve the custom diagnostic task of object part identification (OPI). A key prerequisite to any such analysis is to establish the baseline performance of the VLMs on the task.

Using the method introduced in \S\ref{sec:OPI}, we obtain OPI scores from all models. The scores, averaged across all parts per object (\textit{bird}, \textit{cat}, etc.) and across all objects (\textit{all}), can be seen in \Cref{fig:llava_identify} (left). 
Average object identifiability scores are also included for reference (hatched bars), as induced from part regions.
We observe that the OPI rate is significantly above chance for all three models, at or above 80\% on average, and matches the object identifiability. This shows that the VLMs effectively map between visual concepts and their corresponding vocabulary tokens even on the low level of object parts. While it is expected that object information is often mentioned in text, object parts are mentioned far more rarely.

In \Cref{fig:llava_identify} (right), we see how identifiability progresses across the layers of the models, as a function of percentage of layers computed. OPI in LLaVA 13B progresses in two steps, with one surge at about 20\% and another at 40\% after which it nearly plateaus. LLaVA 7B shows a steep ascent up to about 60\% of its layers, and BakLLaVA's progression is stable throughout. Interestingly, all models show a drop in identifiability in the last 2-3 layers (with a subsequent recovery in the last layer for LLaVA 7B only).
This LLM behavior might be linked to confidence calibration mechanisms, which aim to prioritize a specific set of output tokens, subject to strong language priors in addition to visual context. 

Considering the similarity in trends across the three models, we focus subsequent analysis on one model, LLaVA (13B) for simplicity, presenting partial results for the rest where relevant, with all remaining results included in Appendix \ref{app:additional_results}.

\subsection{Ablating Contextualization}
\label{sec:main_results}

\begin{figure*}[t]
  \centering
  \includegraphics[width=\textwidth]{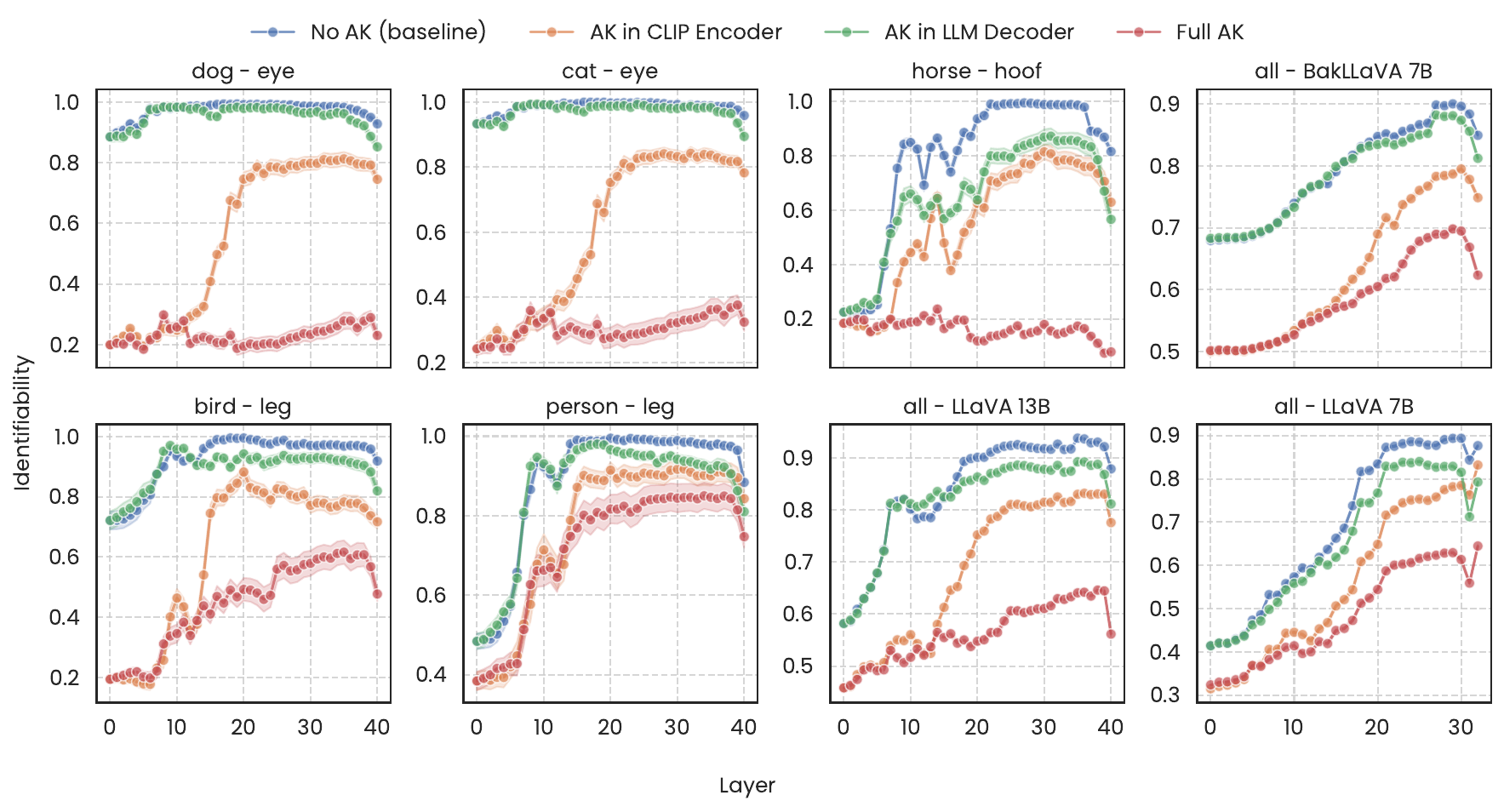}
  \caption{Layer-wise evolution of per-object part identifiability in the LLM Decoder under attention knockout (AK) settings. ``No AK'' denotes unaltered self-attention in both CLIP Encoder and LLM Decoder, while ``Full AK'' indicates modified self-attention in both. The first two columns and top row of the third column show qualitative results for 5 object-part cases for LLaVA 13B model. Remaining plots show aggregated identifiability across all parts for BakLLaVA 7B, LLaVA 7B and LLaVA 13B models.}
\label{fig:llava_attention_knockout_with_llava_13b_qualitative_results}
\end{figure*}

Having established that LLaVA is highly successful in identifying object parts, next we inspect the source of this performance by ablating the contextualization of image patches in the visual encoder and the language decoder through Attention Knockout (AK) \citep{geva2023dissecting}. We block the flow of information between the part regions and the rest of the image, at different layers of LLaVA. This allows us to assess how contextualization plays into OPI across different layers of the encoder and decoder. 

Specifically, we implement AK such that in the encoder,  bidirectional self-attention is blocked between target-region patches and other patches, and in the language decoder, causal self-attention is blocked from target-region to past patches. This is illustrated in Figure~\ref{fig:vlm-attention-map}. 
In both modules, target-region patches can attend to each other, and non-target patches can also attend to each other.

\Cref{fig:llava_13b_attention_knockout} presents the layer-wise identifiability results for all experimental configurations, with a per-object breakdown for LLaVA 13B and averaged results for all three models. 
Below, each configuration is discussed in turn.

\paragraph{Contextualization is Key to OPI.}
To assess the role of contextualization in the OPI task, we establish a floor of performance by applying attention knockout to both the vision encoder and the language decoder (Full AK). This leads to a sizable drop in OPI scores compared to the baseline (No AK), which limited recovery across the layers of the decoder.
Despite this drop, the identifiability rate remains nonzero, indicating that some object parts can be identified to a high degree even in isolation from the object they belong to. In \Cref{fig:llava_attention_knockout_with_llava_13b_qualitative_results}, we show one such case: the leg of a person, whose identifiability drops a bit as a result of the full attention knockout, but stands apart from other parts in how little its identifiability changes. Overall, however, OPI proves to be a highly context-dependent task, as intended.  

Trends are similar for BakLLaVA 7B and LLaVA 7B (last column in \Cref{fig:llava_attention_knockout_with_llava_13b_qualitative_results}), with the gap between the upper (No AK) bound and lower (Full AK) bound being smaller in comparison to LLaVA 13B, but still considerable.

\paragraph{LLM Contextualization Only Marginally Modifies  Good Patch Representations.}
Knocking out self-attention in the decoder results in only slightly lower OPI scores compared to the No AK baseline. This suggests that the self-attention in the decoder plays a small role in the identifiability of object parts. The vast difference in impact between Full AK and this setting 
indicates that most object part information is already encoded in the CLIP encoder via self-attention between target and non-target patches. Therefore, the LLM decoder functions mainly as a ``reader,'' extracting information that is largely self-contained in the visual representations of the target region. 
This is also evidenced by the fact that identifiability rates are relatively high from layer 0 for most object parts, which suggests that the language decoder directly reads much of the semantic information. 

Here again, trends for BakLLaVA and LLaVA 7B are similar, with BakLLaVA in particular exhibiting almost no drop in OPI when self-attention in the decoder is blocked. In light of these observations, the finding of the third and final AK configuration below comes as a surprise.

\paragraph{LLM Contextualization Can Greatly Enhance Poor Patch Representations.}
Based on the previous two AK configurations, we would expect that blocking self-attention in the vision encoder (AK in CLIP Encoder) should have an almost as detrimental effect as blocking all self-attention in the VLM (Full AK), since the role of the LLM appears negligent so far. Yet, we find that OPI scores for this setting greatly exceed floor performance, and actually land much closer to the ceiling performance of the No AK baseline. It appears that the language decoder, while passively reading from strong visual representations, can actively ``write back'' or reconstruct information when presented with weaker visual representations. 

The trend holds similarly for BakLLaVA and LLaVA 7B, albeit with a less pronounced recovery gap for these VLMs with weaker language backbones. Indeed, the size of the LLM appears to factor into how effectively it can compensate for deficiencies in visual representations. The next section further dissects these findings.

\begin{figure}[t]
  \centering
  \includegraphics[width=\columnwidth]{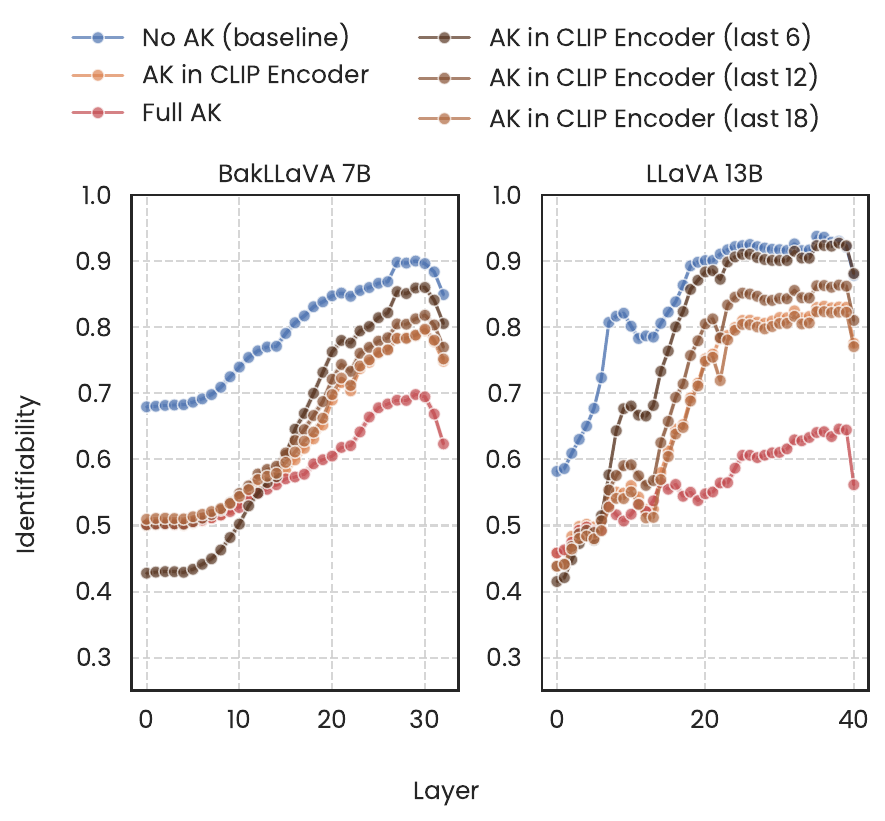}
  \caption{Layer-wise evolution of part identifiability in the LLM Decoder when progressively blocking self-attention across different CLIP Encoder layers of the VLM. ``No AK'' denotes unaltered self-attention in both CLIP Encoder and LLM Decoder, while ``Full AK'' indicates attention knockout in both.}
    \label{fig:progressive_knockout}
\end{figure}

\subsection{How Do LLMs Do It?}
Here, we attempt to discern what type of information the LLM can and cannot recover, and what the mechanism and limiting factors are behind that.

Prior work shows that different layers in vision encoders specialize in different functions, with early layers extracting low-level features and later layers encoding higher-level semantic information \cite{ghiasi2022visiontransformerslearnvisual, dorszewski2025colorsclassesemergenceconcepts}. Motivated by this, we progressively knock out self-attention in the upper layers of the vision encoder--blocking the last 6, 12, 18, or all 24 layers--and measure how well the language decoder can compensate for the lack in visual contextualization. The results are shown in \Cref{fig:progressive_knockout}.

\paragraph{Deep Visual Contextualization is Fully Recoverable (by a strong LLM).}   
When AK is applied to only the last six layers in LLaVA 13B (last 6 in \Cref{fig:progressive_knockout}, right), the identifiability rate remains nearly indistinguishable from the No AK baseline. This finding suggests that the high-level semantic-feature building typically attributed to the last layers of the vision encoder can be fully reconstructed by the LLM. This result is intuitive in light of studies which establish the isotropy of the semantic spaces learned by even independently trained vision and language models \cite{huh2024prh}. Yet, to the best of our knowledge, this is the first time this kind of dynamic division of labor between the modules of a VLM has been observed empirically. 

The potential for full recovery, moreover, appears to be a function of model size. The results with BakLLaVA (\Cref{fig:progressive_knockout}, left) show that a considerable gap remains in OPI scores even at six layers of AK in the CLIP encoder (and similarly for LLaVA 7B in \Cref{fig:llava_7b_clip_progressive_attention_knockout} in the Appendix.) 

\paragraph{Some Part of Shallow Visual Contextualization is Non-recoverable.}
As we extend AK in the vision encoder to 12 and then to 18 layers, we observe that the OPI scores converge towards the numbers seen for full (24-layer) attention knockout. There is almost no difference in what the language decoder can recover from visual representations with six layers of early contextualization with zero layers of early contextualization, meaning that some critical low-level, purely visual feature extraction takes place in those first six layers of the encoder, which is beyond the scope of the LLM decoder. 

Nevertheless, it bears repeating that even with a full 24-layer AK in the CLIP encoder, the LLM is able to compensate for much of the identifiability of object parts -- this was the highlight finding of \S~\ref{sec:main_results}. The analysis above sheds more light on what the LLM cannot compensate for, which appears to be rooted in shallow-layer visual feature extraction. 

\paragraph{The Surge in Identifiability Adaptively Shifts.}
This is best seen in the results for LLaVA 13B, where the no AK baseline exhibits two clear surges, as discussed in \S\ref{sec:baseline}. In \Cref{fig:progressive_knockout} (right), we see that the pattern of the surges is preserved across different levels of attention knockout in the vision encoder. But it shifts from earlier layers to later layers in the decoder, as AK in the vision encoder progresses from the last 6 layers, through 12, 18 and finally 24 layers. 
This delay reflects the compensatory behavior of the LLM decoder: as there is more missing context in the visual representation, the LLM has to dedicate more layers to recover it, effectively ``writing'' key information to the visual representation, before reading it back, i.e., before identifying the object part present in a region.

Lastly, it is worth noting that OPI recovery is observable despite an apparent architectural limitation: language models use a causal self-attention mechanism, meaning that the recontextualization of visual representations in the VLMs' decoders is happening unidirectionally. This may lead to more pronounced shifts in OPI surges, but it does not prevent LLaVA 13B, for example, from achieving full recoverability of visual semantic features.

\begin{figure}[t]
  \centering
    \includegraphics[width=\columnwidth]{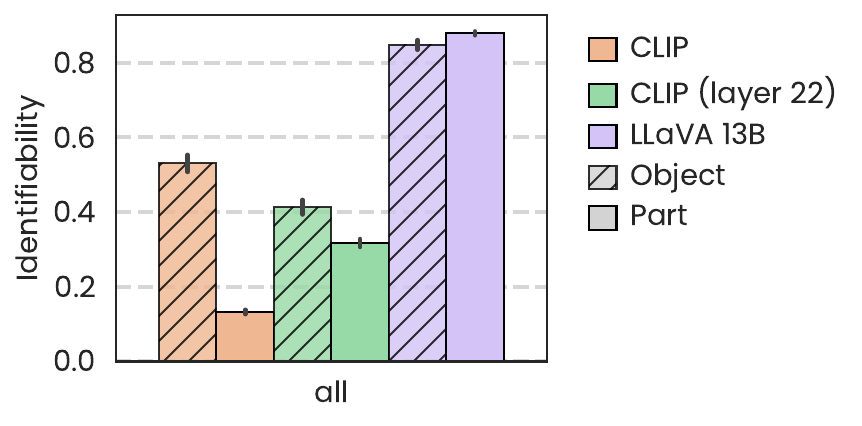}
  \caption{OPI scores for CLIP variants and LLaVA 13B model with scores aggregated across all parts.}
  \label{fig:clip_identify}
\end{figure}

\subsection{Can CLIP Identify Object Parts?}
\label{subsec:clip_identify}

Earlier we saw that different LLaVA variants yield a high OPI score which is largely dependent on the good contextualization of image patches in the CLIP vision encoder. Here, we test whether the CLIP text encoder itself is also able to read this information. Our implementation of the OPI probe is a modification of the standard approach for zero-shot image classification with CLIP.   

\paragraph{Probe Implementation.} 

In zero-shot image classification, CLIP's text encoder is used to obtain representations for all vocabulary tokens contextualized within a template (``A photo of \{token\}''.) and the \texttt{<|endoftext|>} token is used to represent each candidate text. Tokens are then ranked based on their similarity to an input image \texttt{[CLS]} token representation. The standard approach for image classification cannot be used for patch-level analysis off-the-shelf, since the \texttt{[CLS]} token pulls information from all patches in the image. So we implement a simple localization method through self-attention blocking to ensure that the \texttt{[CLS]} token represents a specific image region. Concretely, we allow self-attention to be computed as usual up to a given layer, $l$, in the vision encoder, while for all subsequent layers, self-attention is constrained to the \texttt{[CLS]} token and patches of the target region. 
Unlike the AK interventions used earlier, the procedure described here is not a form of attention ablation, but a method used to focus the representational power of the \texttt{[CLS]} token onto the region of interest within the image. 
We empirically test different values for $l$ and find that $l=22$ works best (see \Cref{fig:clip_attention_knockout_probing_object_parts_wise} in the Appendix). 

\Cref{fig:clip_identify} presents CLIP's OPI scores, including earlier LLaVA 13B results for reference.

\paragraph{CLIP cannot identify object parts.}

The positive impact of the focusing technique on OPI scores is considerable (CLIP vs. CLIP, layer 22), and, as could be expected, zooming in on a particular part region has a negative impact on object identifiability.  
Still, the OPI rates for CLIP (layer 22) are far from the ceiling in performance set by LLaVA. Despite the considerable difference in language model size between CLIP and LLaVA, the gap in their performance more likely stems from deficiencies in the CLIP text encoder, which have been attested in prior work as well \cite{li2024exploring}. The text encoder is not as adept at reading object part information from the image representation, as it is with whole object information. Yet, this does not mean that the relevant information is not present in the CLIP visual encoder, as evidenced by the fact that LLMs can read it from its visual representations. Future work should investigate how CLIP-like models can be so expressive, if they are learned through the bottleneck of pooled representations and with reference to a weak text encoder.


\section{Conclusion \& Future Work}
In this study, we investigated the internal mechanisms of vision-language models (VLMs) by evaluating their ability to identify object parts under a range of controlled attention ablations. Our findings reveal a nuanced and adaptive division of labor between the vision encoder and the language decoder, where an LLM can compensate for some level of deficiency in the visual representations, recovering and enriching semantics that would otherwise be absent. These findings have several implications for future work. First, in more complex visual domains--where even fully contextualized visual features may fall short--the compensatory role of the LLM may become even more critical: LLMs could serve as a fallback mechanism in scenarios where visual representations from the visual encoder alone are sufficiently inexpressive. Second, our results suggest that future research could exploit the capabilities of LLM further for contextualization within VLM architectures. Specifically, disabling self-attention in certain layers of the vision encoder during pretraining--or rethinking the encoder-decoder interface entirely--could lead to more efficient models with deeper fusion of modalities from the ground up. It might be time to rethink VLM architectures: not as static pipelines, but as adaptive systems capable of redistributing computational burden between the vision and language components depending on the task.


\section{Limitations}
While our study provides valuable insights into the contextualization mechanisms of vision-language models, it also has some limitations. We use the Object-Part Identification (OPI) task which, despite no direct applied value, serves as an effective diagnostic probe for isolating and analyzing specific model behaviors in a controlled setting. In this analysis, we rely on a single dataset--Pascal Panoptic Parts which we carefully filter further. We selected this dataset primarily due to detailed, structured annotations of objects and their corresponding parts, which are essential for our task at hand. Most of the other segmentation datasets lack this level of granularity, making them unsuitable for studying contextualization in the way we do in this work.
Furthermore, our study is limited to LLaVA-style models \citep{liu2023visualinstructiontuning,liu2024improvedbaselinesvisualinstruction,zhou2024tinyllava}, which pass all patch representations directly to the LLM decoder; our method does not extend to architectures like BLIP-2 \cite{li2023blip2}, Qwen2-VL \citep{Qwen2VL} or Phi-3.5-Vision\footnote{\url{https://huggingface.co/microsoft/Phi-3.5-vision-instruct}}, which instead rely on sampling or pooling of the original patch representations. We leave the exploration of how our probing framework can be adapted to these newer architectures for future work.

%% file: appendix.tex
\section{Dataset Filtering}
\label{app:dataset_filtering}

In order to maintain annotation consistency and visual clarity, we specifically focus on the subset of animal classes within the Pascal Panoptic Parts (Pascal-PP) dataset \citep{degeus2021panopticparts, meletis2020panopticparts} dataset as described in \Cref{sec:OPI}. The choice is motivated by the relatively uniform and well-defined part structures of animals, as well as their good coverage in the vocabulary of the VLMs we evaluate. To reduce noise and complexity further, we sequentially apply the following filtering criteria:
\begin{enumerate}
    \item Each image contains exactly one instance of the target object (e.g., a single cat). While other non-target objects may be present, multiple instances of the target objects are excluded to avoid ambiguity.
    \item The target object must occupy at least 20\% of the image pixels, ensuring the object is visually prominent enough.
    \item If the target object is fully masked, it must not be mentioned in the caption generated by the VLM to ensure that object and part recognition rely on visible image patches rather than memorization or contextual hallucination.
\end{enumerate}

After filtering, we sample up to 100 images per class, where available. \Cref{tab:dataset_stats} provides further details about the distribution of images across different object classes.

\section{Additional Results}
\label{app:additional_results}

\begin{figure*}[t]
  \centering
  \includegraphics[width=\textwidth]{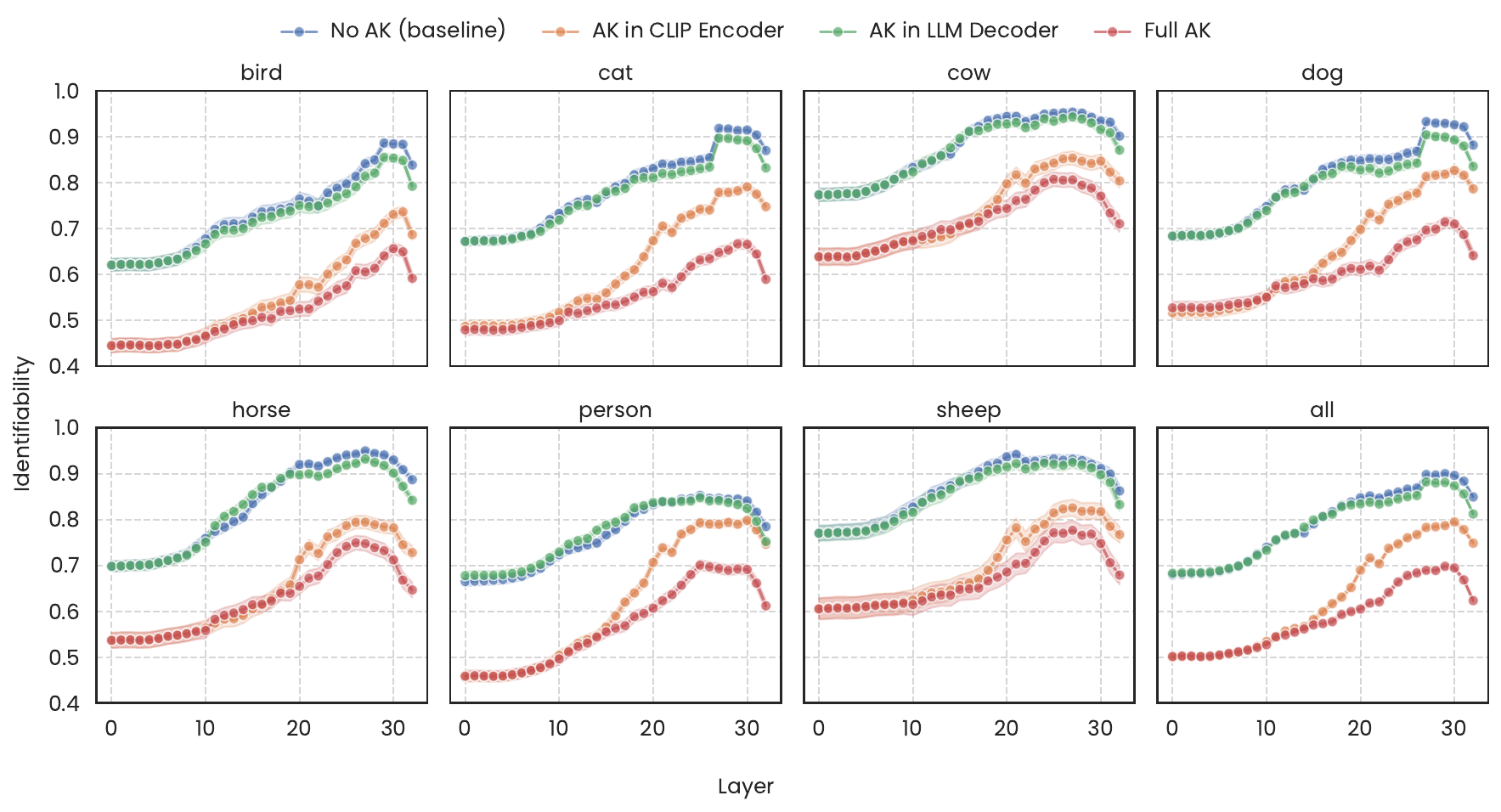}
  \caption{Layer-wise evolution of per-object part identifiability across different attention knockout (AK) settings in the LLM Decoder of the BakLLaVA 7B model. The last plot shows aggregated identifiability scores across all parts. ``No AK'' denotes unaltered self-attention in both CLIP Encoder and LLM Decoder, while ``Full AK'' indicates modified self-attention in both.}
  \label{fig:bakllava_7b_attention_knockout}
\end{figure*}

\begin{figure*}[t]
  \centering
  \includegraphics[width=\textwidth]{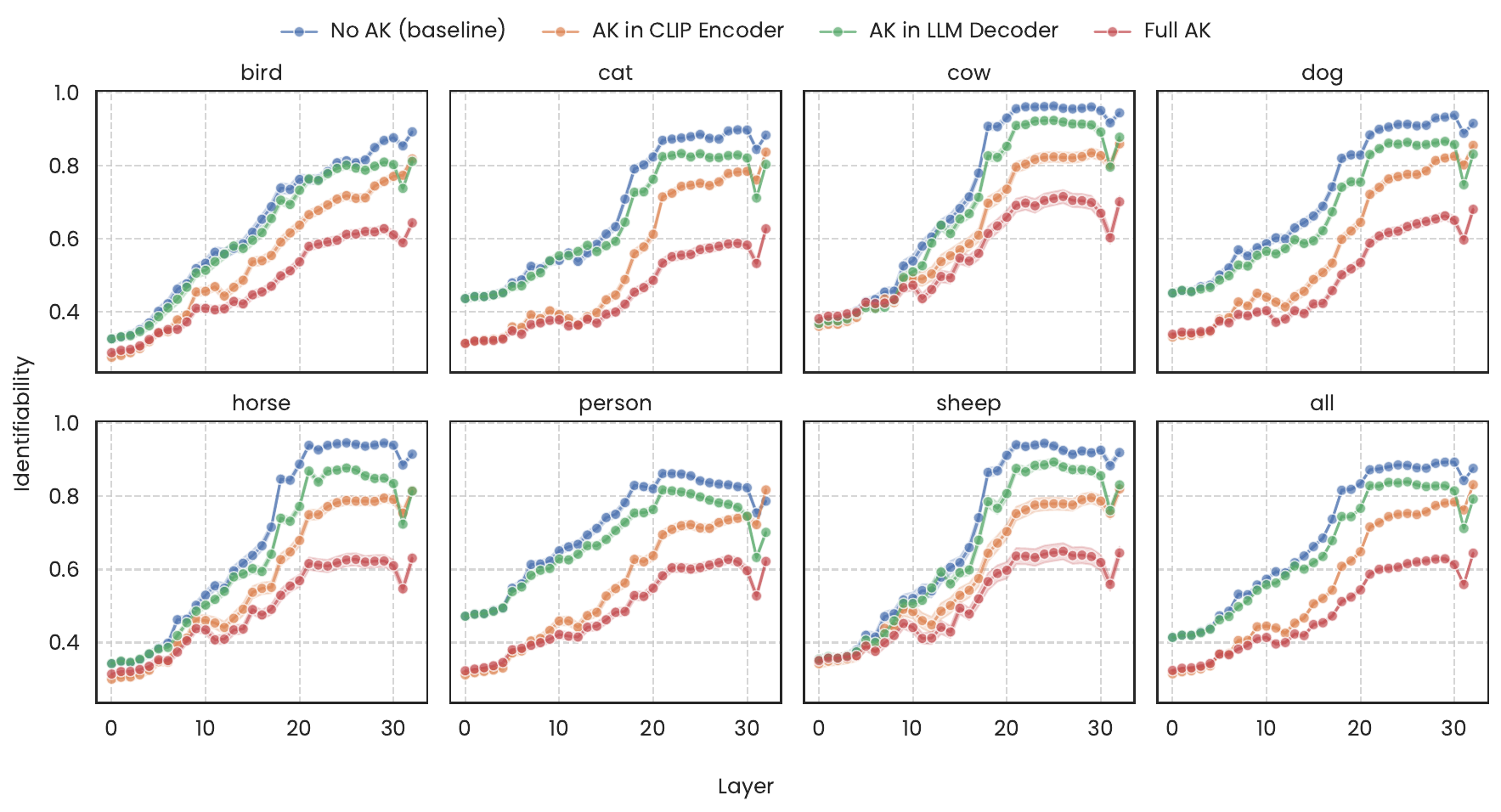}
  \caption{Layer-wise evolution of per-object part identifiability across different attention knockout (AK) settings in the LLM Decoder of the LLaVA 7B model. The last plot shows aggregated identifiability scores across all parts. ``No AK'' denotes unaltered self-attention in both CLIP Encoder and LLM Decoder, while ``Full AK'' indicates modified self-attention in both.}
  \label{fig:llava_7b_attention_knockout}
\end{figure*}

\begin{figure*}[t]
  \centering
  \includegraphics[width=\textwidth]{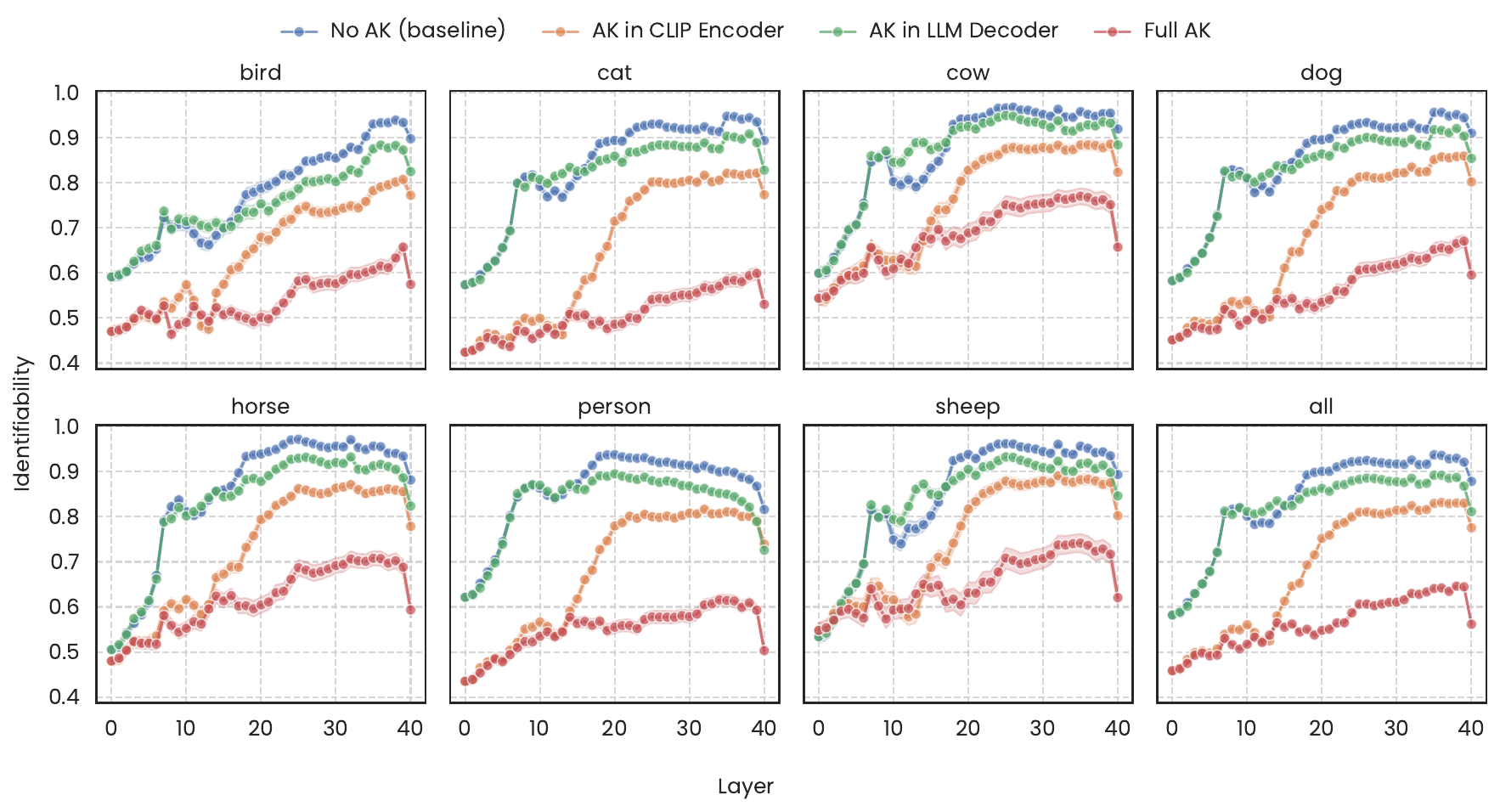}
  \caption{Layer-wise evolution of per-object part identifiability across different attention knockout (AK) settings in the LLM Decoder of the LLaVA 13B model. The last plot shows aggregated identifiability scores across all parts. ``No AK'' denotes unaltered self-attention in both CLIP Encoder and LLM Decoder, while ``Full AK'' indicates modified self-attention in both.}
  \label{fig:llava_13b_attention_knockout}
\end{figure*}

\begin{figure*}[t]
  \centering
    \includegraphics[width=\textwidth]{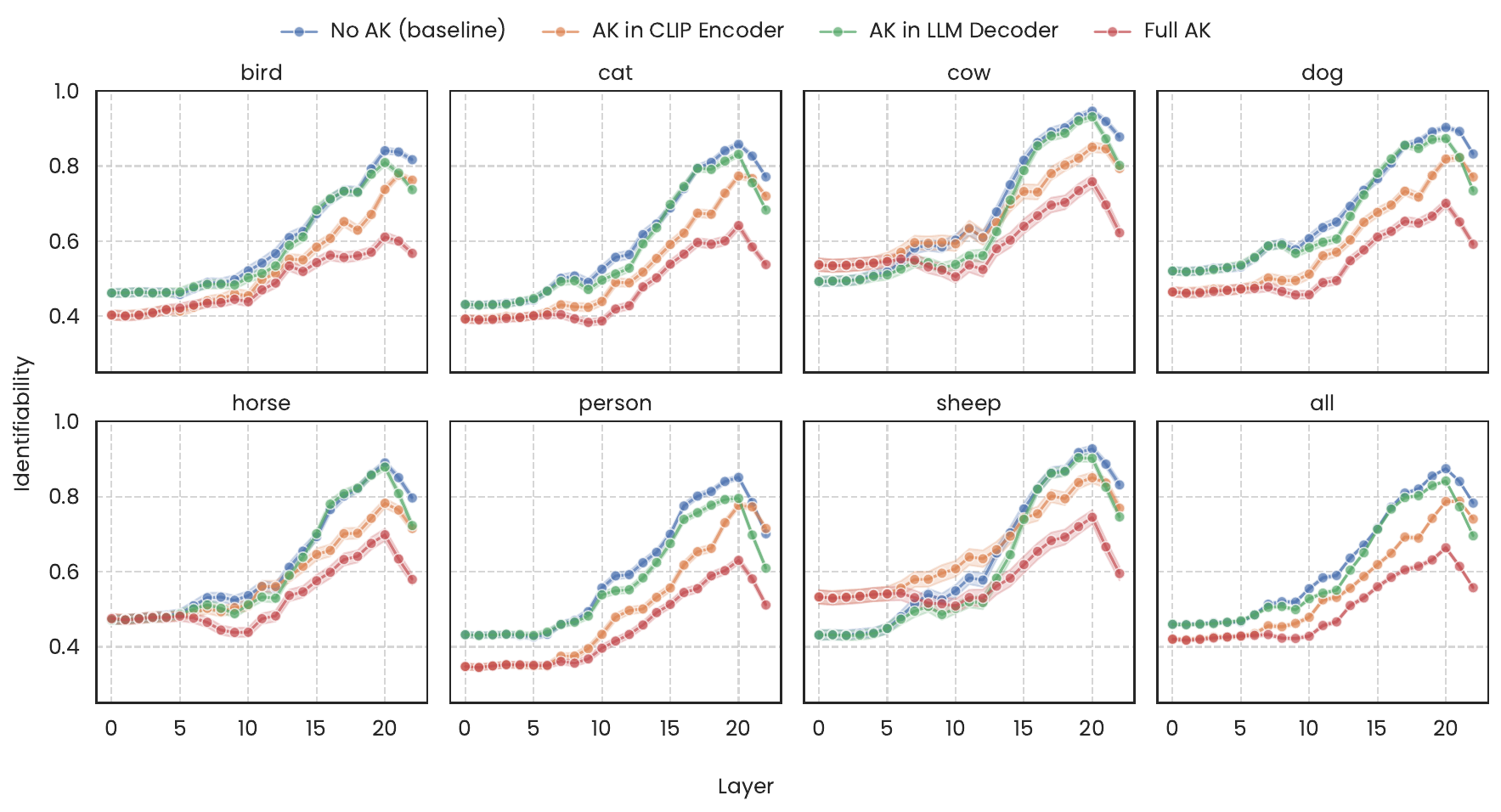}
  \caption{Layer-wise evolution of per-object part identifiability across different attention knockout (AK) settings in the LLM Decoder of the TinyLLaVA 1.1B model. The last plot shows aggregated identifiability scores across all parts. ``No AK'' denotes unaltered self-attention in both CLIP Encoder and LLM Decoder, while ``Full AK'' indicates modified self-attention in both.}
  \label{fig:tinyllava_attention_knockout}
\end{figure*}

\Cref{fig:bakllava_7b_attention_knockout,fig:llava_7b_attention_knockout,fig:llava_13b_attention_knockout,fig:tinyllava_attention_knockout} presents the layer-wise evolution of per-object part identifiability across different attention knockout settings described in the \Cref{sec:main_results} for TinyLLaVA 1.1B, BakLLaVA 7B, LLaVA 7B and LLaVA 13B models.

\begin{figure*}[t]
  \centering
    \includegraphics[width=\textwidth]{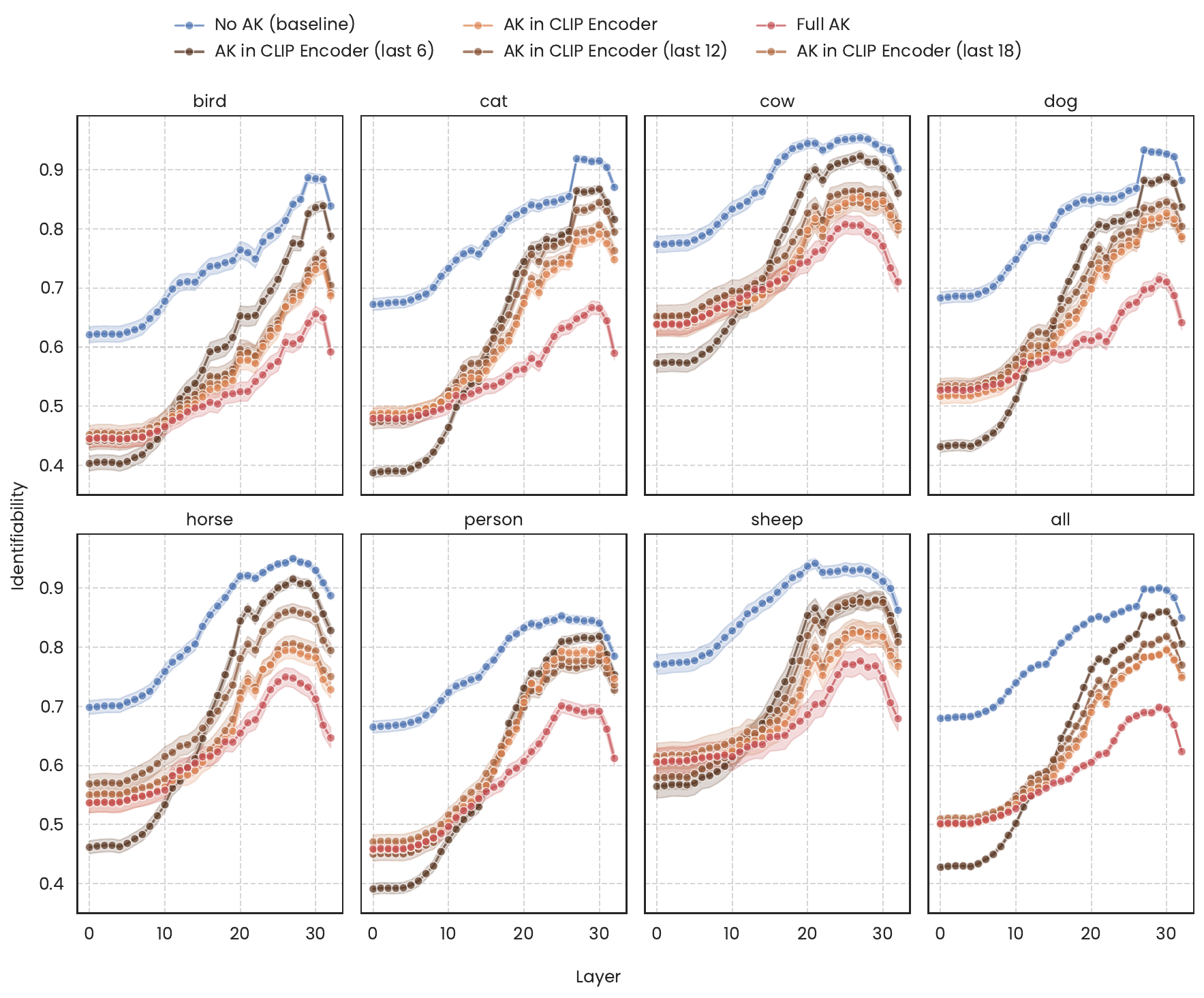}
  \caption{Layer-wise evolution of per-object part identifiability in the LLM Decoder when progressively modifying self-attention across different CLIP Encoder layers in the BakLLaVA 7B model. The last plot presents the aggregated identifiability trends across all objects. ``No AK'' denotes unaltered self-attention in both CLIP Encoder and LLM Decoder, while ``Full AK'' indicates modified self-attention in both.}
  \label{fig:bakllava_7b_clip_progressive_attention_knockout}
\end{figure*}

\begin{figure*}[t]
  \centering
    \includegraphics[width=\textwidth]{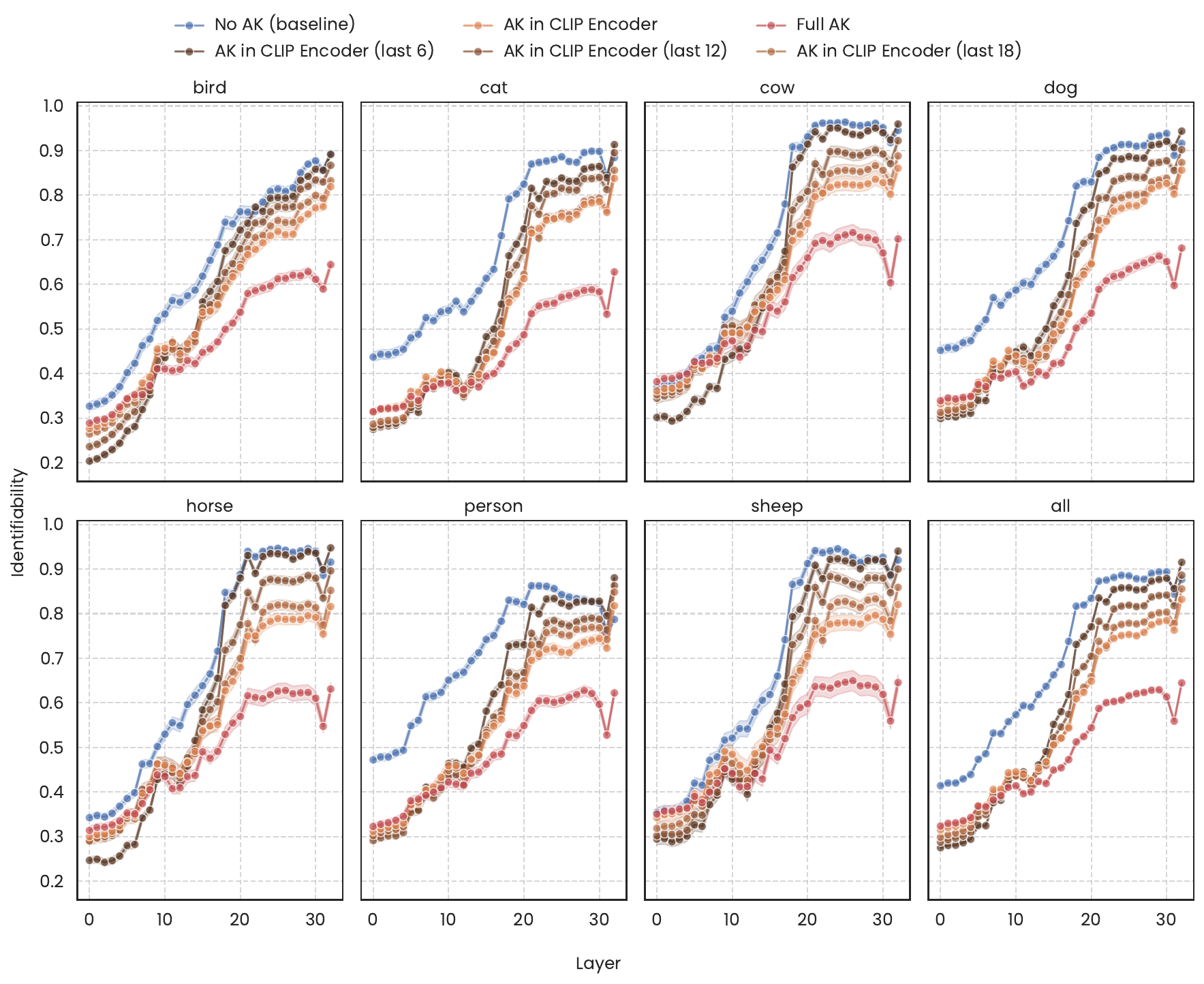}
  \caption{Layer-wise evolution of per-object part identifiability in the LLM Decoder when progressively modifying self-attention across different CLIP Encoder layers in the LLaVA 7B model. The last plot presents the aggregated identifiability trends across all objects. ``No AK'' denotes unaltered self-attention in both CLIP Encoder and LLM Decoder, while ``Full AK'' indicates modified self-attention in both.}
  \label{fig:llava_7b_clip_progressive_attention_knockout}
\end{figure*}

\begin{figure*}[t]
  \centering
    \includegraphics[width=\textwidth]{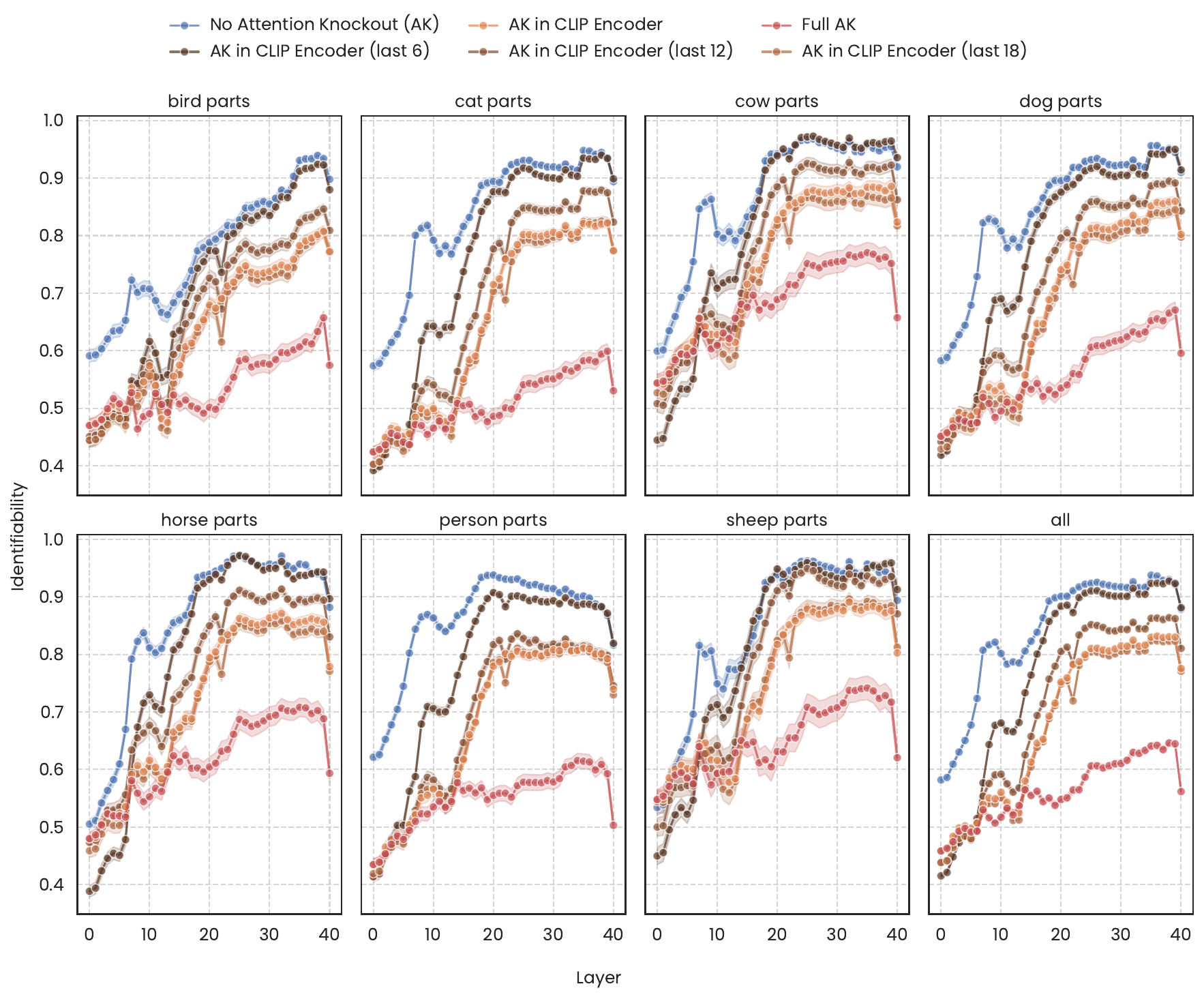}
  \caption{Layer-wise evolution of per-object part identifiability in the LLM Decoder when progressively modifying self-attention across different CLIP Encoder layers in the LLaVA 13B model. The last plot presents the aggregated identifiability trends across all objects. ``No AK'' denotes unaltered self-attention in both CLIP Encoder and LLM Decoder, while ``Full AK'' indicates modified self-attention in both.}
  \label{fig:llava_13b_clip_progressive_attention_knockout}
\end{figure*}

\Cref{fig:bakllava_7b_clip_progressive_attention_knockout,fig:llava_13b_clip_progressive_attention_knockout,fig:llava_7b_clip_progressive_attention_knockout} presents the layer-wise evolution of per-object part identifiability when progressively modifying self-attention across different layers of the CLIP Encoder for BakLLaVA 7B, LLaVA 7B and LLaVA 13B models.

\begin{figure*}[t]
  \centering
  \includegraphics[width=0.8\textwidth]{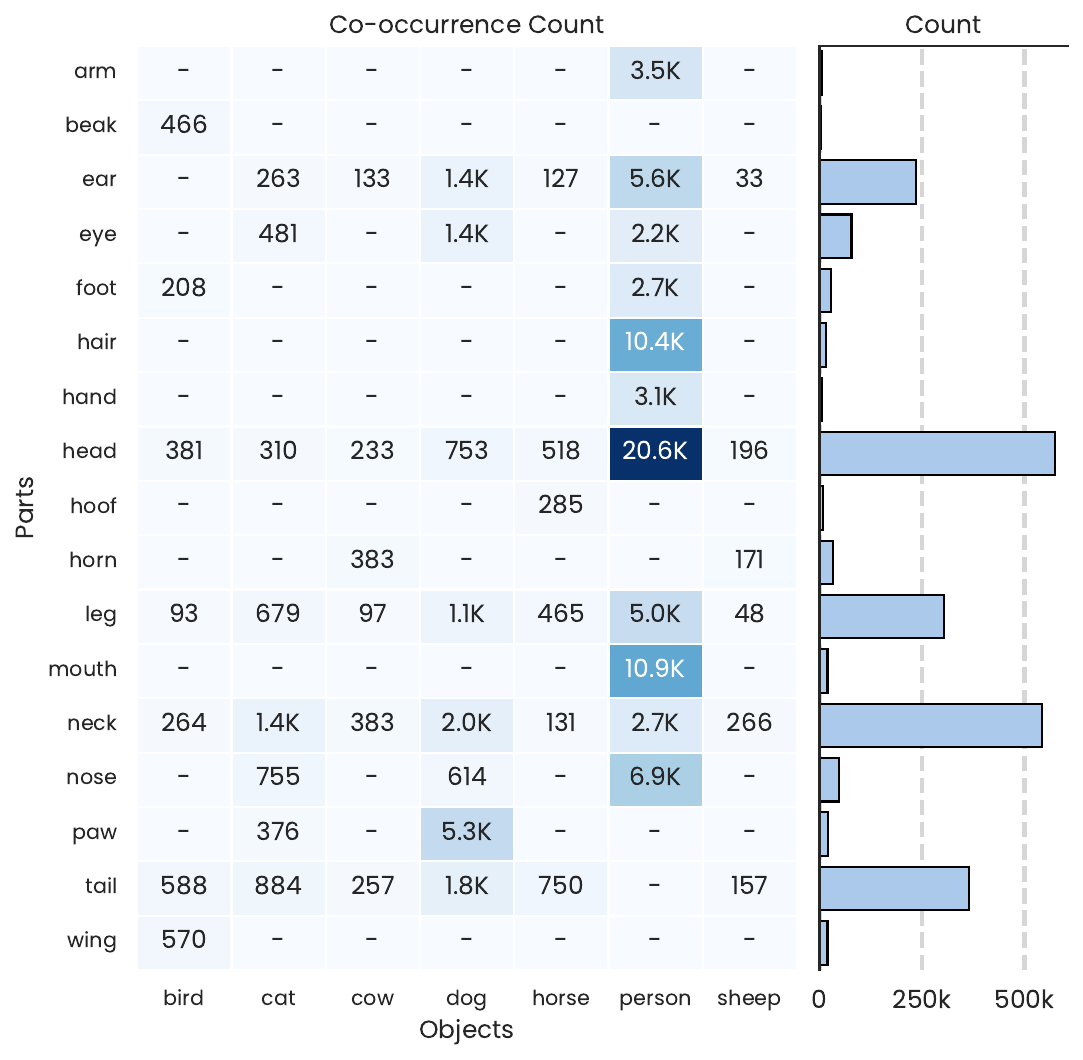}
  \caption{Statistics of frequency counts on the LLaVA dataset from both the pre-training alignment and instruction fine-tuning phase. The heatmap illustrates the frequency of object-part co-occurrence with the associated objects in the context, whereas the barplot shows the frequency of counts of object-parts irrespective of the associated objects.}
  \label{fig:part_count_and_co_occurrence_count}
\end{figure*}

\section{Relationship between identifiability and frequency of object-parts}

In this section, we examine whether the OPI scores correlate with the frequency of object-part co-occurrences or just parts irrespective of the associated objects in the training data of the models we evaluate on. All the models employ a two-stage training process where the first stage focuses on alignment using the image captioning data and the second stage focuses on instruction fine-tuning using the multimodal instruction data. We combine the data instances from both stages and compute the co-occurrence counts of objects and their respective parts in the ground-truth responses using the following pipeline:

\paragraph{Text Normalization.} We first lowercase all responses and remove punctuation. Next, each word is stemmed using the Porter Stemmer to match morphological variants (e.g., "legs" vs. ``leg'').

\paragraph{Lexical Expansion.} Each object term and its corresponding parts are expanded by collecting synonyms and hyponyms from WordNet. This expansion of the target set allows for broader lexical coverage and more precise matching of candidate terms. For consistency, we also apply stemming to the terms in the expanded set.

\paragraph{Matching.} We iterate through all instances and match the tokens with the target set of all object terms, part terms, and also keep track of their co-occurrence.

The heatmap in \Cref{fig:part_count_and_co_occurrence_count} presents the co-occurrence counts between different objects (columns) and their associated parts (rows), whereas the bar plot displays the overall frequency of individual parts irrespective of object association. We can observe that co-occurrence frequencies vary widely across object-part pairs. Common parts such as \textit{head}, \textit{leg}, and \textit{eye} frequently appear with the \textit{person} class, whereas parts like \textit{hoof} and \textit{wing} are more sparsely represented, mainly occurring with specific objects like \textit{horse} and \textit{bird} respectively. One might hypothesize that object parts associated with frequently occurring classes -- particularly \textit{person} would exhibit higher identifiability. However, analysis of OPI scores (see \Cref{tab:llava_variants_opi_scores}) reveals no clear correlation between co-occurrence frequency and identifiability. In fact, parts with relatively low frequency often achieve high identifiability. Similarly, in \Cref{fig:llava_attention_knockout_with_llava_13b_qualitative_results}, parts such as \textit{eye} in the context of \textit{dog} and \textit{cat}, and \textit{leg} in the context of \textit{bird} and \textit{person} are identified accurately despite their lower co-occurrence counts. These findings suggest that the identifiability of object parts in VLMs is not predominantly driven by training data frequency. Instead, VLMs are capable of implicitly learning object-part associations leveraging strong visual or semantic priors, and exhibit effective generalization capabilities, even without explicitly being trained for fine-grained object parts recognition.

\begin{figure}[t]
  \centering
  \includegraphics[width=\columnwidth]{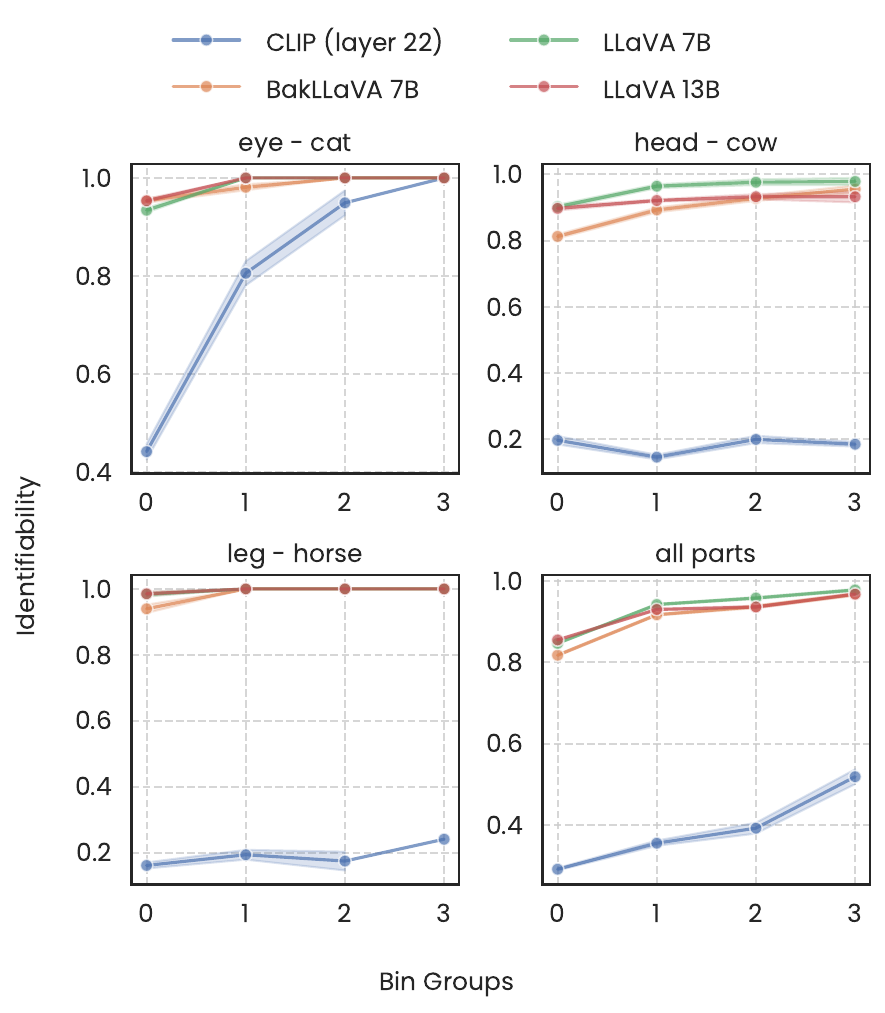}
  \caption{Qualitative results illustrating the relationship between OPI scores and part size across bin groups for CLIP model (best-performing configuration) and three LLaVA variants: BakLLaVA 7B, LLaVA 7B, and LLaVA 13B models. The last plot shows aggregated OPI scores across all object parts.}
  \label{fig:identify_vs_parts_size}
\end{figure}

\section{Relationship between identifiability rate and size of object parts}

In this section, we investigate whether the identifiability of object parts is influenced by their relative size. Intuitively, one might expect larger parts to be easier to identify due to their increased visibility and spatial prominence in images. To test this, we consider the CLIP model (with best-performing configuration $l=22$; see \Cref{subsec:clip_identify}) and three LLaVA variants: BakLLaVA 7B, LLaVA 7B, and LLaVA 13B. We begin by binning object parts sizes into four discrete groups based on their relative area in the image space. We then analyze the identifiability rates of object parts across these bin groups. \Cref{fig:identify_vs_parts_size} presents identifiability scores for few object-part pairs such as \textit{eye - cat}, \textit{head - cow}, and \textit{leg - horse}, as well as aggregate trends over all parts.

Our analysis highlights a key difference between the CLIP and LLaVA models. For CLIP, identifiability scores generally increase with part size for several part-object combinations such as \textit{eye - cat}, \textit{leg - horse}, and \textit{paw - dog}, indicating a positive correlation between part size and identifiability. However, for certain parts like \textit{head - cow}, identifiability remains consistent across bin groups. In contrast, the three LLaVA variants show little to no variation in identifiability with respect to part size. Across all bin groups, a consistent gap in identifiability exists between CLIP model and LLaVA variants, though the gap narrows as the part size increases. These results demonstrate that while part size moderately influences identifiability in CLIP, it has minimal impact on the LLaVA variants, suggesting insensitivity to part size in current VLMs.

\section{Potential Application of LLaVA for Segmentation Task}

In this section, we test the extent to which LLaVA’s strong object and part recognition capabilities translate to the task of object parts segmentation. We utilize the OVParts benchmark \citep{wei2023ovpartsopenvocabularysegmentation} to evaluate segmentation performance across both LLaVA 7B and 13B models. Specifically, we assess the models' ability to perform part segmentation over a subset of 850 images from the OVParts validation set. To ensure fair comparison, we follow LLaVA’s original preprocessing pipeline across all samples. Unlike traditional segmentation models, which are trained with pixel-level supervision, LLaVA was not trained for segmentation and instead operates at a patch-based resolution. To adapt LLaVA for segmentation, we map the token with the highest ranking to all pixels in its associated patch. While this approach may produce coarse segmentations and introduce misalignments, it allows us to quantify LLaVA’s segmentation performance in an open-vocabulary setting without additional training.

For comparison, we benchmark against CLIPSeg \citep{lueddecke22_cvpr}, a model fine-tuned for segmentation and among the top-performing entries on the OVParts leaderboard. Unlike LLaVA models, CLIPSeg benefits from supervision explicitly tailored to the segmentation task, allowing it to generate high-fidelity pixel-wise masks. We report the mean IoU (mIoU) scores  \citep{mIoU} in \Cref{tab:segmentation_results}. We find that LLaVA 7B and 13B models achieve comparable mIoU scores, but they underperform the CLIPSeg baseline considerably. The performance gap is expected given LLaVA’s lack of segmentation-specific training and the mismatch between patch-level predictions and pixel-level evaluation metrics. However, the better-than-random mIoU scores again attest to LLaVA's ability to identify object parts and suggest that the model could be a good starting point for dedicated segmentation training.

\begin{table}[t]
    \centering
    \begin{tabular}{lc}
        \toprule
        Model & mIoU \\
        \midrule
        CLIPSeg (baseline) & \textbf{22.31} \\
        LLaVA 7B & 14.66 \\
        LLaVA 13B & 14.81 \\
        \bottomrule
    \end{tabular}
    \caption{Comparison of mIoU scores on OVParts part segmentation between LLaVA 7B/13B models and CLIPSeg specifically tuned for segmentation.}
    \label{tab:segmentation_results}
\end{table}

\begin{figure*}[t]
  \centering
  \includegraphics[width=0.9\textwidth]{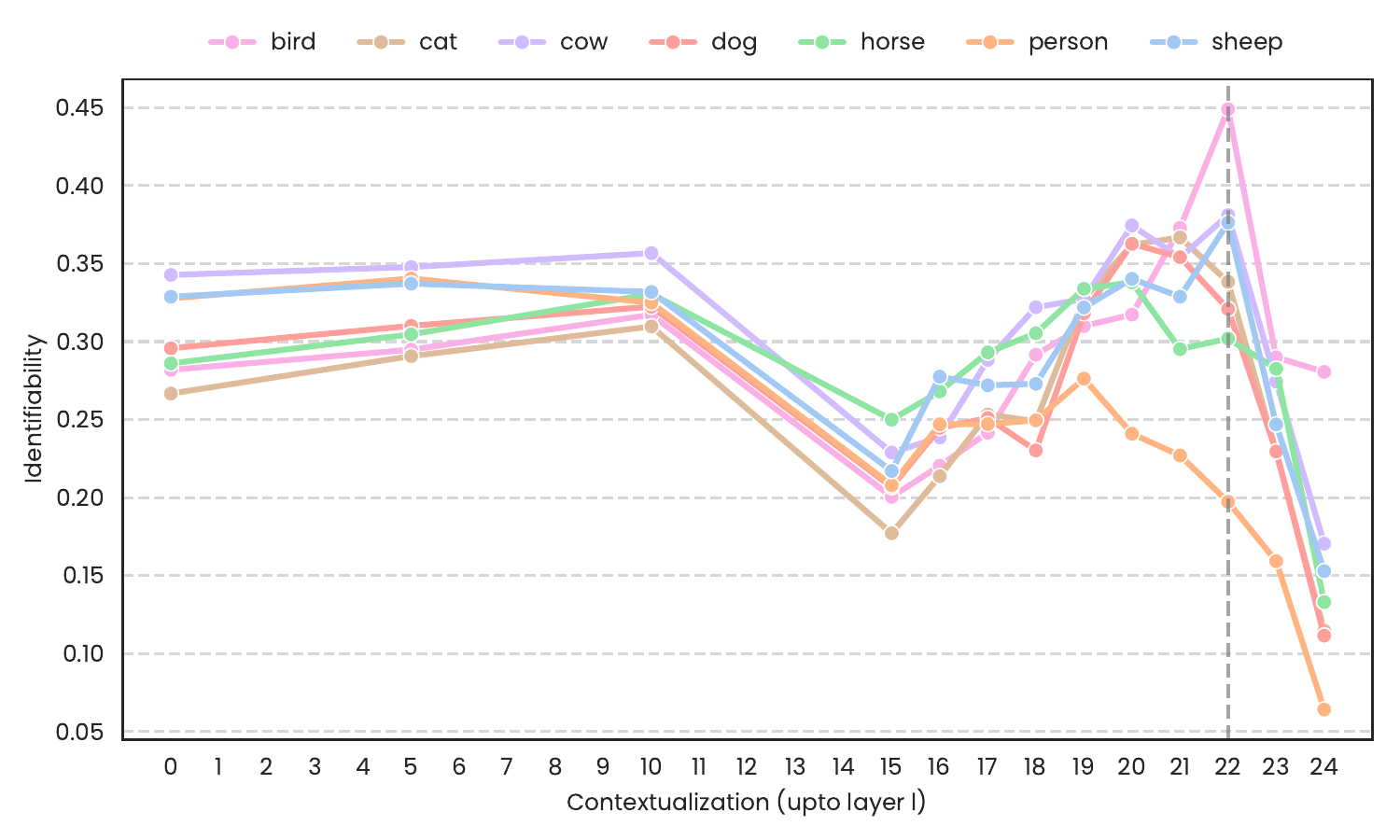}
  \caption{Effect of layer-wise localization in the CLIP Encoder on per-object part identifiability using attention knockout.}
\label{fig:clip_attention_knockout_probing_object_parts_wise}
\end{figure*}

\begin{table*}[]
\centering
\begin{tabular}{lcccccccc}
\toprule
\multirow{2}{*}{Object Category}  & \multicolumn{2}{c}{BakLLaVA 7B} & \multicolumn{2}{c}{LLaVA 7B} & \multicolumn{2}{c}{LLaVA 13B} & \multicolumn{2}{c}{TinyLLaVA 1.1B} \\
       & part & object & part & object & part & object & part & object \\
\midrule
bird   & 0.84 & 0.83   & 0.89 & 0.83   & 0.90 & 0.85   & 0.82 & 0.90   \\
cat    & 0.87 & 0.86   & 0.88 & 0.88   & 0.89 & 0.89   & 0.74 & 0.87   \\
cow    & 0.90 & 0.92   & 0.95 & 0.95   & 0.92 & 0.93   & 0.88 & 0.94   \\
dog    & 0.88 & 0.90   & 0.92 & 0.91   & 0.91 & 0.91   & 0.72 & 0.93   \\
horse  & 0.89 & 0.90   & 0.92 & 0.91   & 0.88 & 0.94   & 0.80 & 0.94   \\
person & 0.78 & 0.51   & 0.79 & 0.57   & 0.82 & 0.52   & 0.67 & 0.45   \\
sheep  & 0.86 & 0.96   & 0.92 & 0.96   & 0.89 & 0.94   & 0.83 & 0.93  \\
\bottomrule
\end{tabular}
\caption{OPI scores for BakLLaVA 7B, LLaVA 7B, LLaVA 13B and TinyLLaVA 1.1B models, showing part-level and object-level identifiability aggregated across all parts.}
\label{tab:llava_variants_opi_scores}
\end{table*}

